# Group-Skeleton-Based Human Action Recognition in Complex Events

Tingtian Li
Interactive Entertainment Group
Tencent Inc.
Shanghai, China
tingtianli@tencent.com

Zixun Sun
Interactive Entertainment Group
Tencent Inc.
Shanghai, China
zixunsun@tencent.com

Xiao Chen
Interactive Entertainment Group
Tencent Inc.
Shanghai, China
evelynxchen@tencent.com

## ABSTRACT

Human action recognition as an important application of computer vision has been studied for decades. Among various approaches, skeleton-based methods recently attract increasing attention due to their robust and superior performance. However, existing skeleton-based methods ignore the potential action relationships between different persons, while the action of a person is highly likely to be impacted by another person especially in complex events. In this paper, we propose a novel group-skeleton-based human action recognition method in complex events. This method first utilizes multi-scale spatial-temporal graph convolutional networks (MS-G3Ds) to extract skeleton features from multiple persons. In addition to the traditional key point coordinates, we also input the key point speed values to the networks for better performance. Then we use multilayer perceptrons (MLPs) to embed the distance values between the reference person and other persons into the extracted features. Lastly, all the features are fed into another MS-G3D for feature fusion and classification. For avoiding class imbalance problems, the networks are trained with a focal loss. The proposed algorithm is also our solution for the Large-scale Human-centric Video Analysis in Complex Events Challenge. Results on the HiEve dataset show that our method can give superior performance compared to other state-of-the-art methods.

## 1 INTRODUCTION

Human action recognition is an active research area and plays an important role in many applications such as surveillance [1], video content analysis [2], and human-computer interaction [3]. Earlier methods for action recognition depend on hand-crafted features [4-7]. [4, 5] first detect sparse feature points then trace their temporary trajectories for action classification. However, these sparse key points cannot cover all meaningful regions. [6, 7] improve their performance via sampling dense features and use optical flows to describe the trajectories. Although the dense features can cover all the regions in the video, the deficiency of hand-crafted features still lowers their performance. With the development of deep neural networks, deep-learning-based methods become popular [8-12]. [8] and [9] specially designed two-stream convolutional neural network (CNN) architectures to extract spatio-temporal features from an RGB reference image and another optical flow sequence. The final action is determined by the combination of output scores of the two networks. However, the spatial features only extracted from the reference image may lose details in other frames and the optical flow sequence also easily suffers from the occlusion problem when movements of a person are large. For conquering this problem, [10, 11, 13] try to directly learn features from image sequences. [10, 12] directly learn spatio-temporal features by feeding a deep 3-dimensional convolutional network (C3D) with an image sequence. However, convolving a sequence of images can consume large memories, which limits their flexibility in the design of the network architectures. [11, 13] use Long Short-Term Memory (LSTM) networks to learn the representation of video sequences. But the direct use of RGB frames still introduce interferences from the various backgrounds and person appearances, which can be noisy for the action classification.

Recently, skeleton-based action recognition methods attract increasing attention from researchers. Compared to conventional RGB-based methods, the skeleton-based human action recognition methods often show more robust performance. The skeleton data can be represented as a sequence of body key point coordinates. The assembly of these compact data can avoid the interference from the complex backgrounds and is more robust to viewpoint changes and different person appearances. Although people can treat the skeleton as a pseudo-image and feed it to recurrent neural networks (RNNs) [14, 15] or CNNs [16, 17], the most efficient way to exploit skeleton data is using Graph Convolutional Networks (GCNs) [18-22]. GCNs can efficiently access the irregular skeleton key points and extract their features in the spatio-temporal domain [18]. [18] is the first work that develops spatio-temporal GCN for action recognition and achieves superior performance. [19-22] further investigate GCNs and improve the performance with specifically designed architectures [19, 21, 22] or using Neural Architecture Search (NAS) [20]. However, none of them considers the potential action relationship between different persons in a video. Indeed, actions of people staying close can have strong relationships, especially in complex events. For example, a person very close to another person who is queuing for a long time has a high probability of also being queuing; a person near another person who is fighting is highly likely to act as the fighting opponent. Such a phenomenon is not a coincidence, we also can find many other actions e.g. gathering, talking, walking, etc. have this property. In this paper, we exploit this hint and propose a novel GCN-based approach that utilizes group-skeleton data with distance embedding. In addition to conventional key point coordinates, we also input the key point speed values to the networks. For better exploiting skeleton features in multiple scales, we adopt the MS-G3D [19] as the graph feature extraction network. Furthermore, for avoiding the possible class imbalance problem, the networks are trained using a focal loss [23]. Results on the HiEve dataset show that the proposed method can give superior performance compared to other state-of-the-art approaches.

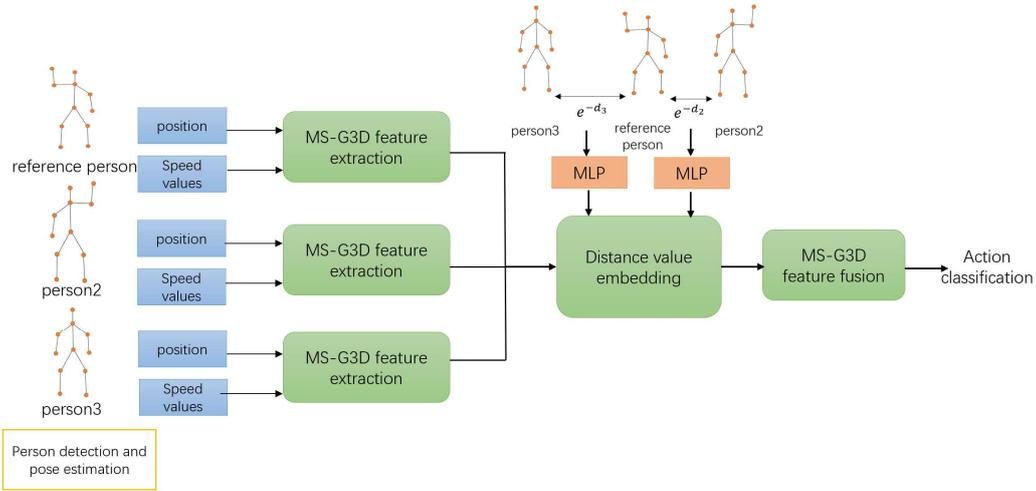

Figure 1: The flow chart of the proposed algorithm. Here, we input three persons as an example.

## 2 THE PROPOSED METHOD

Figure 1 shows the flow chart of the proposed algorithm. We first detect the persons in the video and estimate their poses. Then we input their key point positions and speed values to MS-G3D for feature extraction. As closer persons should have a stronger action relationship, we also embed the distance values between the reference person and other persons to the extracted features. All the features will be fused by another MS-G3D. Finally, the classification result is output by a fully connected layer. As far as we know, we are the first one that combines skeleton data of multiple persons with the GCN for action recognition. In the rest of this section, we will present the details of our algorithm.

### 2.1 Action Recognition Using Group-Skeleton Data

Conventional GCN based approaches only consider the skeleton sequence of a target person. However, as mentioned before, the actions of other persons around the target person are also highly likely to impact each other. Therefore, we propose to extract the features of other $M-1$ persons who are nearest to the reference person. $M$ is the number of persons whose features are extracted. The skeleton sequence can be obtained by pose estimation algorithms [24-27] and pose tracking techniques [27-29]. Considering the pose speed can be a useful hint to distinguish some actions like standing and walking, we also feed the networks with pose speed values. The speed values can be obtained by

$$v_i^k(t) = p_i^k(t) - p_i^k(t-d), \quad (1)$$

where $p_i^k$ is the coordinate of the $i$th key point of the $k$th person, $t$ represents the frame index, $d$ is the frame interval for calculating the key point speed, $v_i^k$ is the speed value for the $i$th key point of the $k$th person. Here, we denote $k = 0$ for the reference person. In case that some key points move back to original position when the time interval is long, we adopt a small value 3 for $d$. For every person, we concatenate $p_i^k(t)$ and $v_i^k(t)$ for all valid $i$ and $t$ as a tensor $z^k$ with dimensions $(C, T, K)$, where $T$ and $K$ are the frame and key point number respectively, $C = 3$ contains key point coordinates $(x, y)$ and a confidence value $c$. For every person, we input this tensor to a MS-G3D for feature extraction. This process can be represented as $f^k = G^k(z^k)$, where $G^k$ represents the feature extraction network for the $k$th person. Here, we use MS-G3D for feature extraction is because MS-G3D can better exploit key point movements at different scales and easily learn an adaptive graph with residual masks [19]. The output feature dimensions of $f^k$ for the $k$th person are $(C_1, T/2, K)$. We can find the time dimension is halved. This is because the MS-G3D temporarily down-samples the features.

## 2.2 Distance Embedding Using MLPs

After skeleton features of each person are obtained, the intuitive way to fuse them is directly concatenating them as a tensor and input the tensor to another network. However, the direct concatenation cannot reflect the observations that close persons should have strong action relationships while distant persons may have mild or no action relationships. Therefore, we propose to use MLPs to embed the distance values into the extracted features. The MLPs are used to align the distance features with the extracted features and can also boost the effectiveness of the distance features via learning. The embedding process can be represented as

$$\hat{f}^k = \begin{cases} f^k \cdot mlp^k(e^{-d^k}) & if\ k \geq 1 \\ f^k & if\ k = 0 \end{cases}, \quad (2)$$

where $\hat{f}^k$ is the distance embedded feature of the $k$th person, $mlp^k$ is an MLP for the $k$th person, $d^k$ is the pose distance between the $k$th person and the reference person, the symbol '·' means element-wise multiplication. We can find the input $e^{-d^k}$ drops when the distance increases. For the reference person, as the distance to itself is zero, we define its feature does not change. $d^k$ is a tensor with dimensions $(1, T/2, K)$. It stacks distance values between the corresponding key points of two persons every two frames. $d^k$ not only gives the distance of two persons but also measures the relative movements of the corresponding key points of the two persons. Because close persons tend to make the same actions, the movements of the corresponding key points should have some correlations. The MLP can be represented as follows

$$mlp^k(x) = \sigma\left(BN(W^k x + b^k)\right), \quad (3)$$

where $W^k \in \mathbb{R}^{C_1 \times 1}$, $b^k$ is a bias vector, $BN$ is a batch normalization layer, $\sigma$ denotes the ReLU function. Notice the MLP can align the distance values with $f^k$. The dimensions of $mlp^k(e^{-d^k})$ are $(C_1, T/2, K)$.

## 2.3 Feature Fusion and Action Classification

After distance value embedding, we fuse the features and classify the action for the reference person. Here, we use another MS-G3D to fuse the features. We concatenate all the embedded features to be a tensor $[\hat{f}^0, \hat{f}^1, ..., \hat{f}^{M-1}]$ and input this tensor to the MS-G3D for feature fusion. The fused features will be sent to a fully connected layer followed by a softmax function for classification. The outputs of the softmax function can be considered as the probabilities of the different classes. Considering some actions e.g. fighting, fall-over are rare compared to other common actions e.g. standing, walking, we adopt the focal loss [23] to address the class imbalance problem. The focal loss can be expressed as

$$FL = \sum_j a_j (1 - Prob_j)^\gamma log(Prob_j), \quad (4)$$

where $Prob_j$ is the result of the softmax function for the $j$th class, $\gamma = 2$ is a balance factor, $a_j = 1$ when the $j$th action is the ground truth, otherwise $a_j = 0$. Compared to the cross-entropy loss, the focal loss can assign a larger $(1 - Prob_j)^\gamma$ to $log(Prob_j)$ when $Prob_j$ is small. Thus, an action imbalanced class can have a higher weight, which results in a larger gradient for correct convergence.

## 2.4 Network Architectures

All the MS-G3Ds at the first feature extraction stage have the same architecture. The architecture contains two basic STGC blocks [19], both with 8 spatial scales ranging from 1 to 8 and two temporary sliding windows with size 3 and 5. The temporary stride size of the first STGC is 1, and the second one is 2. The output channel numbers of the first and second STGC blocks are 96 and 192 respectively. The MS-G3D used for feature fusion only has one STGC block with 8 spatial scales ranging from 1 to 8 and two temporary sliding windows with size 3 and 5. Its stride size is 2 and the output channel size is 384.

## 3 Experiments and Results

### 3.1 Implementation Details

In this subsection, we will give the implementation details of our algorithm. The algorithm is also our solution for the Large-scale Human-centric Video Analysis in Complex Events Challenge [30], where we ranked the 4th place. Their HiEve dataset contains more than 56k complex-event action labels of 14 action classes. As we find most complex events in the HiEve dataset can be represented by at least three persons, e.g. queuing, gathering, we set the input person number $M$ to be 3. In this challenge, we use EffientDet [31], CrowdPose [25] and PoseFlow [28] to detect persons, estimate and track person poses respectively. We use SGD to train the networks with batch size 16, weight decay 0.0005, and an initial learning rate 0.05. The learning rate decays with a factor 0.1 when the epoch number reaches 100, 200, 300, and 400 epochs. We stop the training process when we find the loss changes very slightly.

As the ground truth for the HiEve test dataset is not public, for developing and validating the effectiveness of components of our algorithm, we partition the training dataset into 4:1. The first four-fifths are used to train the network and the last fifth is used to evaluate the necessity of each component of the network. Notice we only use the second part of the dataset to evaluate the network after the complete of the whole training process, therefore, it can be treated as a 'model-evaluation' dataset. In the final stage of the competition, we use all the data in the training dataset to train our network and produce the results of the trained model on the challenge test dataset. The results are then submitted to the challenge website for evaluation. The evaluation result will be output from the website according to their specific matrices.

Table 1: The model component comparison results on the 'model-evaluation' dataset

| Method | Accuracy (%) |
|---|---|
| GS-GCN w/o multiple persons (MS-G3D) | 71.62 |
| GS-GCN w/o pose speed values | 73.71 |
| GS-GCN w/o the distance embedding | 73.03 |
| GS-GCN | **76.26** |

Table 2: Results of different methods on the HiEve test dataset

| Method | wf-mAP@avg | wf-mAP@50 | wf-mAP@60 | wf-mAP@75 | f-mAP@avg | f-mAP@50 | f-mAP@60 | f-mAP@75 |
|---|---|---|---|---|---|---|---|---|
| RPN+I3D | 6.88 | 9.65 | 7.91 | 3.07 | 8.31 | 11.01 | 9.65 | 4.26 |
| Faster R-CNN+I3D | 10.13 | 13.35 | 11.57 | 5.49 | 10.95 | 14.50 | 12.33 | 6.01 |
| Transformer+I3D | 7.28 | 9.88 | 8.32 | 3.65 | 7.03 | 9.32 | 8.10 | 3.66 |
| GS-GCN | **15.09** | **20.30** | **17.31** | **7.66** | **16.25** | **21.44** | **18.77** | **8.55** |

### 3.2 Components of The Model Architecture

For validating the effectiveness of each component of the model architecture, we conduct an ablation study on the 'model-evaluation' dataset. For convenience, we name the proposed method as GS-GCN. For validating the use of features of multi-person is essential, we compare this method but only extracting only one person. Notice, in this case, the network degrades to the MS-G3D but still with speed values as the input. For demonstrating the distance embedding is necessary, we also show the results of the GS-GCN without using distance embedding. For verifying the speed values are a useful hint, we also compare this method only using the conventional key point coordinates as the input. The ablation study results are shown in Table 1. As this ablation study aims at comparing the performance of this method using different architecture components without considering the pose estimator and tracker used in our framework, we only use the accuracy as the evaluation matrix. We discard those detected persons with intersection over union (IoU) value below 0.3 in respect of the ground truths for sustaining the least quality of the inputs and define one ground truth bounding box mostly corresponds to only one detected person who has the largest IoU. From Table 1, we can find with all the components, the GS-GCN can give the best performance.

### 3.3 Comparisons

We also compare our whole framework including the pose estimator and tracker with other state-of-the-art methods on the HiEve test dataset. These compared methods are RPN+I3D [32], Faster R-CNN+I3D [30, 33], and Transformer+I3D [34]. They are all implemented and evaluated on the HiEve test dataset by the HiEve team [30]. RPN+I3D is a strong baseline for AVA challenge [32]. It uses I3D [35] to extract features and feed the features to RPN [33] for region proposal and action classification. The faster R-CNN+I3D is an improved version of RPN+I3D which applies a faster R-CNN detector on the keyframes and obtains the person bounding boxes as action classification proposals [30]. Transformer+I3D [34] also uses I3D as its backbone but it adds an attention mechanism that lets the network focus on crucial parts of human bodies. The comparison results are shown in Table 2. To evaluate the performance, we use the frame mean Average Precision (f-mAP) and the weighted frame mAP (wf-mAP) [30] that are used in the Challenge to measure the performance. f-mAP is a common metric to evaluate spatial action detection accuracy. wf-mAP is a weighted version of f-mAP. It assigns a smaller weight to dominant action classes and a larger weight to frames under crowded and occluded scenarios for handling the unbalanced distribution of action classes in the data set. As the weights are not public, we obtained the measurement results by uploading our results to the challenge website during the competition. The comparison results are shown in Table 2, where we can find the GS-GCN can outperform other start-of-the-art methods by large margins. This is owing to the use of compact skeleton data of multiple persons and the efficient GCNs for feature extraction, while other methods using RGB data of a single person cannot exploit the potential action relationships of multiple persons and also suffer from interferences from the various backgrounds and person appearances that lower their performance.

## 4 CONCLUSIONS

In this paper, we propose a novel GCN-based algorithm named GS-GCN for action recognition in complex events. This algorithm is also our solution for the Large-scale Human-centric Video Analysis in Complex Events Challenge. Unlike conventional methods that only consider the actions of individual persons, the proposed method investigates the potential action relationships between different persons. We use multiple MS-G3Ds to simultaneously extract skeleton features from multiple persons. As closer persons can have stronger action relationships, MLPs are used to embed the distance values to the extracted features. After a feature fusion step, we train the networks with a focal loss to classify different actions. To our knowledge, we are the first one that combines group-skeleton data with the GCN for action recognition. Experimental results on the HiEve dataset show that our method can achieve superior performance compared to other state-of-the-art methods.